\documentclass[10pt,twocolumn,letterpaper]{article}
\pdfoutput=1
\usepackage{cvpr}
\usepackage{times}
\usepackage{graphicx}
\usepackage{amsmath}
\usepackage{amssymb}
\usepackage{float} 
\usepackage{booktabs}
\usepackage{multirow}
\usepackage{algorithm}
\usepackage{algpseudocode}
\usepackage{chngpage}
\usepackage[normalem]{ulem}

\usepackage[pagebackref=true,breaklinks=true,letterpaper=true,colorlinks,bookmarks=false]{hyperref}

\cvprfinalcopy % *** Uncomment this line for the final submission

\begin{document}

%%%%%%%%% TITLE
\title{Decoupled Appearance and Motion Learning for Efficient Anomaly Detection in Surveillance Video}

% \author[1]{Bo \snm{Li}\corref{cor1}} 
% \cortext[cor1]{Corresponding author:  Tel.: +32 9 264 33 16;}
% \ead{Bo.Li@UGent.be}
% \author[1]{Sam \snm{Leroux}}
% \author[1]{Pieter \snm{Simoens}}

% \address[1]{IDLab, Department of Information Technology, Ghent University - imec, Ghent 9000, Belgium}

\author{Bo Li \quad Sam Leroux \quad Pieter Simoens\\
IDLab, Department of Information Technology, Ghent University - imec,\\
Ghent 9000, Belgium\\
{\tt\small Bo.Li@UGent.be}
% \and
}

\maketitle
%\thispagestyle{empty}

%%%%%%%%% ABSTRACT
\begin{abstract}
Automating the analysis of surveillance video footage is of great interest when urban environments or industrial sites are monitored by a large number of cameras. As anomalies are often context-specific, it is hard to predefine events of interest and collect labelled training data. A purely unsupervised approach for automated anomaly detection is much more suitable. For every camera, a separate algorithm could then be deployed that learns over time a baseline model of appearance and motion related features of the objects within the camera viewport. Anything that deviates from this baseline is flagged as an anomaly for further analysis downstream. We propose a new neural network architecture that learns the normal behavior in a purely unsupervised fashion. In contrast to previous work, we use latent code predictions as our anomaly metric. We show that this outperforms reconstruction-based and frame prediction-based methods on different benchmark datasets both in terms of accuracy and robustness against changing lighting and weather conditions. By decoupling an appearance and a motion model, our model can also process 16 to 45 times more frames per second than related approaches which makes our model suitable for deploying on the camera itself or on other edge devices.
\end{abstract}

%%%%%%%%% BODY TEXT\section{Introduction}
\section{Introduction}
Rising concerns for public security and safety have increased the number of surveillance cameras installed in our streets and public places~\cite{abati2019latent, liu2018ano_pred, Ionescu_2019_CVPR, Chandola:2009:ADS:1541880.1541882, morais2019learning}. 
Human operators in a control room continuously inspect these video streams on a multi-screen video wall, looking for abnormal events that may mandate further inspection. As human operators can only process a few video streams at the same time, part of the surveillance workflow must be automated when the number of cameras grows. By automating the detection of anomalous events, human operators can focus on the appropriate response to these events, e.g.~by requesting a police intervention.

Since anomalies are context-specific~\cite{song2007conditional}, each video stream requires a tailored anomaly detection algorithm. A running person for example is considered an anomaly in a busy shopping street but it might be normal in a train station as people are often in a hurry to catch the train. In this work we introduce a new neural network architecture that is able to recognize anomalous events in a surveillance camera stream using only unsupervised training. We propose to decouple the learning of appearance and motion information which are the key factors for determining anomalies in a surveillance video. We first train an autoencoder by reconstructing individual frames to capture high level appearance features such as the location, shape and size of an object. Such features however can not guarantee the detection of  anomalies that are caused by motion related features such as speed or trajectory. We therefore add a second component that further exploits the spatiotemporal information of the frequently seen events by predicting the latent code for a future frame using the stacked latent codes of the previous $k$ frames as the input. The underlying assumption is that the anomalous events are rare occasions and will not be modeled accurately by the networks. The predicted latent code of anomalous frames will hence deviate significantly from the observed latent codes.

% We propose a decoupled architecture consisting of two neural networks. First, we train an autoencoder to extract latent codes from individual frames. Then, we use the codes extracted from the previous $k$ frames to predict the latent code of a future frame. The underlying hypothesis is that anomalous events are rare occasions and will not be modelled accurately by the networks. The predicted latent code of anomalous frames will hence deviate significantly from the extracted latent code. 

Our approach is easy to implement and achieves state-of-the-art performance on benchmark datasets. We however do not only focus on detection accuracy but also address several other obstacles for real-world deployment. Our model is much more efficient than related approaches, which could make it possible to evaluate our model at the network edge, on or nearby the surveillance camera itself, as opposed to streaming all data to a central point for analysis. Inference at the edge is also a more privacy friendly paradigm since a human operator will not see the camera data unless his intervention is needed. Lastly, our experimental results indicate that detection performance based on prediction of latent codes is more robust against changing weather and lighting conditions.

The remainder of this paper is organized as follows: In section~\ref{section:related} we give an overview of related anomaly detection methods. In section~\ref{sec:method} we introduce our approach and we experimentally validate it on different benchmark datasets in section~\ref{sec:experiments}. In section~\ref{sec:robustness_of_the_model} we show that our approach is more robust against different distortions. We conduct an ablation study in section~\ref{sec:ablation_study} to analyze the role of different components of the model. We conclude in section~\ref{sec:conclusion} and give a few pointers for future research directions. 
\section{Related work}
\label{section:related}
% Anomaly detection is a useful technique for many use cases such as intrusion detection in computer networks or fault detection in industrial appliances~\cite{Chandola:2009:ADS:1541880.1541882}.
% Different techniques are available depending on the type of input data. 

Deep learning is currently the state-of-the-art method for many computer vision related tasks~\cite{SCHMIDHUBER201585} and is also the technique behind the state-of-the-art anomaly detection methods for video surveillance type data. We can differentiate three different approaches to do anomaly detection with deep learning: reconstruction based methods, prediction based methods and methods that use characteristics of the latent code to detect anomalies.

%\subsection{Deep Learning Based Anomaly Detection}
%In surveillance videos, the anomalies are usually caused by motion or insertion of a foreign object or instrumentation errors~\cite{Chandola:2009:ADS:1541880.1541882}, so we consider the related anomaly detection methods from reconstruction, prediction and latent code aspects.

%\textbf{Reconstruction based methods}
\subsection{Reconstruction based methods}
%Deep learning methods are driving advances in many computer vision related tasks~\cite{SCHMIDHUBER201585} as well as anomaly detection~\cite{RIBEIRO201813} with a powerful feature representation capacity. 
The most common approach is to build models that reconstruct their input. These models are based on an autoencoder architecture that contains a bottleneck for encoding high level features, creating a compressed representation of the input data. These compressed representations are then used to reconstruct the input data. The assumption here is that the reconstruction works fine for inputs that are similar to the data that was seen during training but that anomalous inputs can not be modelled accurately by the learned features, resulting in a poor reconstruction. The reconstruction error is then used as a metric to detect anomalies.

For our use case of anomaly detection in video surveillance, it is not enough to only model spatial information by processing individual frames, we also need to consider the temporal information to detect anomalies that are caused by motion, such as high speed or irregular movement. Different approaches have been explored to incorporate this information into the model.~\cite{Hasan2016LearningTR} exploit the spatiotemporal information by reconstructing multiple stacked frames using an autoencoder.~\cite{XU2017117} apply a denoising autoencoder~\cite{Vincent:2008:ECR:1390156.1390294} to reconstruct frames. They use optical flow maps to describe the motion information. Another option is to use Recurrent Neural Networks (RNN) or Long Short-Term Memory networks (LSTM)~\cite{SCHMIDHUBER201585} to model the time dimension~\cite{10.1007/978-3-319-59081-3_23, 8019325}. Finally, there is also work that explores the use of a 3D convolutional networks (C3D) to model spatiotemporal representations.~\cite{Tran:2015:LSF:2919332.2919929} show that C3D can encapsulate information related to shapes and motions in video sequence better than a 2D based model, thus boosting the anomaly detection accuracy. %All of these methods assume that the spatiotemporal information can be employed by reconstructing the stacked frames and the model will assign a high reconstruction error to the abnormal events, and these assumptions do not necessarily hold. As a consequence, either the motion-related anomaly is hardly to be recognized or the normal and abnormal events have similar reconstruction error.

\subsection{Prediction based methods}
Instead of reconstructing the input, it is also possible to predict future frames. This requires a better understanding of temporal information.~\cite{liu2018ano_pred} use a generative adversarial network (GAN)~\cite{NIPS2014_5423} that takes stacked frames and optical flow features as input. Anomalies are detected at test time by measuring the difference between the predicted and observed future frame. Any deviation from the expected frame is considered as an anomaly. For more specialized applications we can also use domain specific features.~\cite{morais2019learning} for example deal with human-related anomaly detection by reconstructing and predicting the decomposed global body movement and local body posture from the human skeleton movement. %Differently from the above mentioned works, our method i) works on the single frame instead of stacked frames so that each frame only need to be processed once ii) predict the features for the future frame thus alleviate the blurry prediction problem of using pixel-wise Mean Square Error (MSE) as objective function~\cite{Mathieu2015DeepMV} iii) does not require any preprocessing step such as the computation of optical flow maps in order to apply motion constraints on anomaly detection. 

\subsection{Latent code based methods}
Both lines of previous work generate expected frames and detect anomalies by measuring the difference in pixel space with the actual input frame. However, it is well known that pixel-wise similarity measurements do not necessarily correspond with human understanding of images~\cite{Larsen:2016:ABP:3045390.3045555, Mathieu2015DeepMV} and are often very sensitive to minor changes in brightness or color. On the other hand, the high level features extracted by a neural network are shown to be less sensitive to these distortions~\cite{Zhang2018TheUE}. There are some very recent approaches that extract high level features with an autoencoder and then use a classifier such as a one-class SVM on the extracted features~\cite{app9040757, Ionescu_2019_CVPR} to detect anomalies. The assumption is that the classifier will distribute the anomalies outside of the learned manifold. The work that is most similar to our approach is the Latent Space Autoregression model from~\cite{abati2019latent}. They use features from a deep 3D convolutional autoencoder combined with an autoregressive network to model the probability distribution underlying the latent representation. They combine the reconstruction error and the likelihood of the latent code to identify the anomalies. Our approach is similar in that we also extract features and work in a high level latent code. However, our approach uses a 2D autoencoder to extract features which highly reduces the number of parameters and computational cost. We use a less complicated feed forward model to do the prediction which does not make any assumptions on the distribution family of the latent code. It allows us to directly use the Mean Squared Error (MSE) between the predicted latent code and the latent code that is extracted from the encoder as the anomaly score. Finally, we explicitly predict the latent code of a future frame instead of relying on the reconstruction of the current frame to capture the motion information.

\begin{figure*}
\centering
\resizebox{.9\textwidth}{!}
{
\begin{minipage}{\textwidth}
\includegraphics[width = 1.0\textwidth]{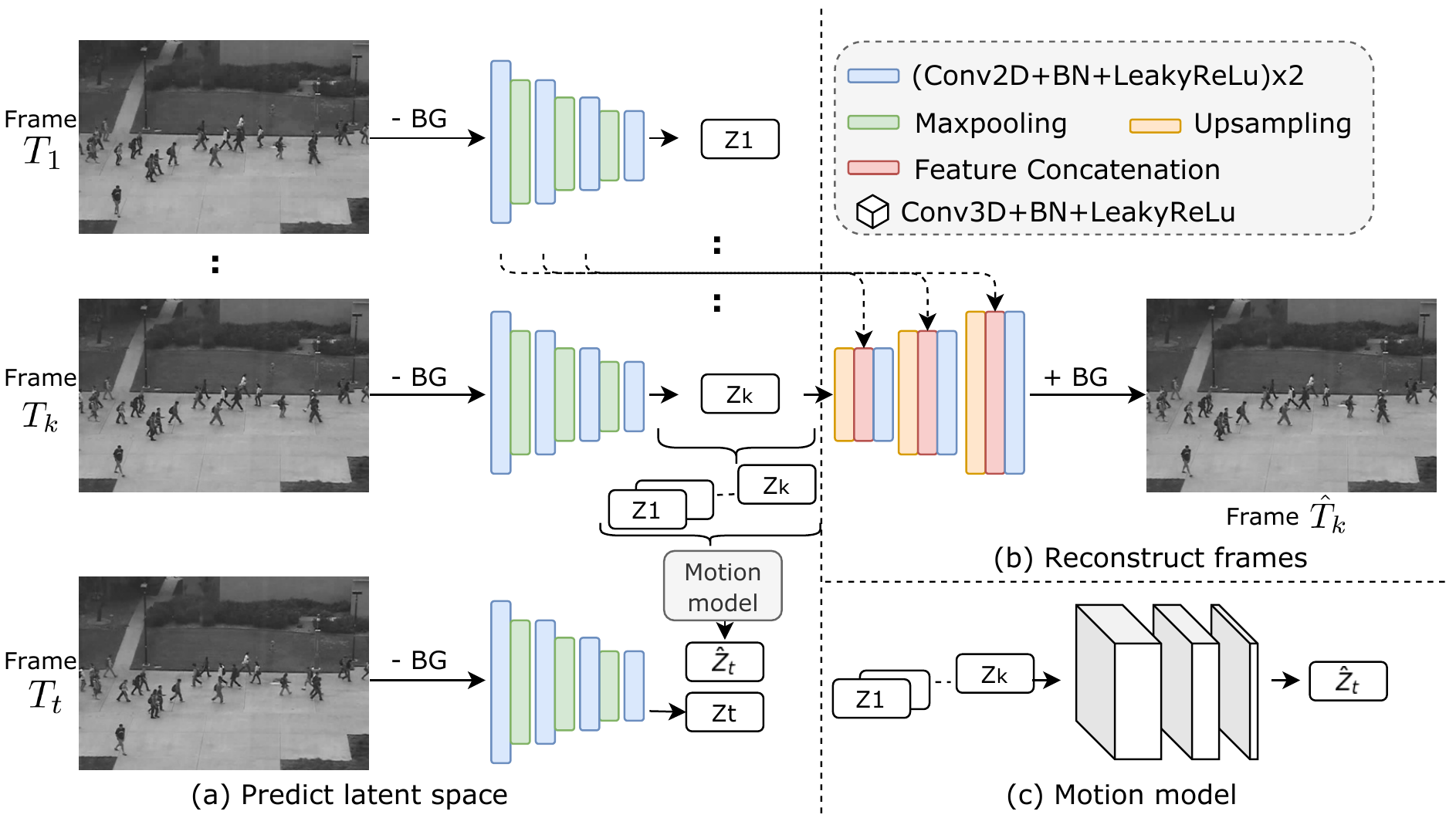}
\end{minipage}
}
\caption{Overview of our approach. We use the same encoder to extract latent code for each input frame, where $k$ is the number of frames in per input sequence and $t$ is the frame that is 6 timesteps into future ($k+6$). In the training phase, these latent code are used to (a) predict latent codes for future frames with the motion model and (b) reconstruct current frames with the decoder. In the inference phase, we only use the encoder and the motion model. (c) \textit{Conv3D} layers are used in the motion model to learn spatiotemporal information. The figure is best viewed in color.}
\label{fig:main_architecture}
\end{figure*}
\section{Architecture}
\label{sec:method}

In this paper we propose a decoupled architecture to learn the spatiotemporal information which is important for determining anomalies in surveillance videos. We first train an autoencoder to reconstruct individual input frames and aim to represent the appearance information such as shape, location and outlook of an object in the latent codes. Then to further stress the appearance information for frequently seen events and the dynamical aspects in a video, we stack these extracted latent codes from the encoder for a sequence of $k$ frames and use that as the input for a second network to predict the latent code for a future frame. The model is assumed to be only able to predict the latent codes for frequently seen events with high accuracy. The difference between the predicted and the observed latent codes is then used as the anomaly metric. The following sections explain these in detail.

% First we extract latent codes of individual frames using an autoencoder that is trained to reconstruct individual input frames. Then, we use the encoder part of this autoencoder to extract latent codes for a sequence of $k$ frames. We stack these latent codes and use this as the input for a second network that predicts the latent code for a future frame. The difference between the predicted and the observed latent code is then used as a metric to decide how anomalous the frame is. The following sections explain both parts in more detail. 

\subsection{Learning appearance features}
To learn high level features, we use a U-Net~\cite{RFB15a} type autoencoder that is trained to reconstruct individual input frames. To force the model to focus on the foreground, we subtract a background frame from each input frame. This background frame is calculated as a frame with per-pixel mean RGB values over all training data. The encoder learns to extract latent codes from a single frame and the decoder learns to reconstruct the input based on the extracted features. The original U-Net architecture has shortcut connections between encoder and decoder. To avoid the trivial solution of copying feature maps from the encoder to the decoder and to improve the regularization power, we add a shortcut connection between the previous frame $T_{k-1}$ and current frame $T_{k}$. In other words, the feature maps that are calculated using frame $T_{k-1}$ are concatenated with the feature maps from frame $T_{k}$ in the upsampling path for reconstructing frame $T_{k}$. The detailed architecture is shown in Figure \ref{fig:main_architecture} (a) and (b).

\subsection{Learning motion features}
Spatiotemporal information is important for detecting anomalies in videos. However, the features that are extracted by encoding a single frame as described above can only focus on the spatial information such as shape, location and size of an object and cannot guarantee the detection of the motion-related anomalies. We thus include a second component that can attend to both spatial and temporal dimension to further learn the dynamical aspects of video sequence. Previous work considered learning the temporal information either by predicting optical flows using a pretrained FlowNet~\cite{Hasan2016LearningTR,liu2018ano_pred} or by predicting future frames in pixel space~\cite{abati2019latent, liu2018ano_pred}. This however has three drawbacks. First, the optical flow estimation is computationaly expensive as it requires around $0.1$ seconds to evaluate a single frame on a GPU machine~\cite{IMKDB17}. Secondly, we need to consider the interaction between appearance and motion information. For example, a vehicle driving with very high speed is usually an anomaly except when it is an ambulance. Independently encoding the appearance and motion information using a pretrained optical flow model cannot take this into account. Finally, the pixel-wise Mean Square Error (MSE) objective function for predicting future often generates blurry frames~\cite{Mathieu2015DeepMV, Zhang2018TheUE}. Instead, we decide to predict the latent code of a future frame through a small motion learning model as shown in Fig.~\ref{fig:main_architecture} (c).

% The encoding of single frame input cannot guarantee the detection of motion-related anomalies. Spatiotemporal information is an important factor in videos, which is why we include a second component that can further learn the dynamic aspects of a video sequence. Previous work considered learning the temporal information either by predicting optical flow maps using FlowNet~\cite{Hasan2016LearningTR, liu2018ano_pred} or by predicting future frames in pixel space~\cite{abati2019latent, liu2018ano_pred}. This however has two main drawbacks. First, the optical flow estimation is computationally expensive as it requires around $0.1$ second to evaluate a single frame on a GPU~\cite{IMKDB17}. Second, the pixel-wise Mean Square Error (MSE) objective function often generates blurry frames~\cite{Mathieu2015DeepMV, Zhang2018TheUE}. Thus, we decide to  predict the latent code of a future frame through a small motion learning model as shown in Figure \ref{fig:main_architecture} (c). 

To predict the latent code of frame $T_t$, we extract latent codes for $k$ previous input frames $T_1$...$T_{k}$. The extracted latent codes $z_{1}...z_{k}$ from the encoder for each of past frames $T_1...T_{k}$ are then concatenated along the temporal dimension and used as the input for the motion model to predict the latent code $\hat{z}_t$ for a future frame $T_t$. We use 3D convolutional layers in the motion model since these can attend to both motion and appearance whereas a 2D convolution layer is only able to work in the spatial direction~\cite{Tran:2015:LSF:2919332.2919929}. Each convolutional block includes a 3D convolutional layer with kernel size 3x3x3, stride 2 on the temporal dimension and stride 1 on the feature dimension. This is followed by a BatchNormalization~\cite{Ioffe:2015:BNA:3045118.3045167} and a leaky-relu activation layer. We use three convolutional blocks in the motion model.

%The motion model has four convolution blocks, each with a 3D convolutional layer with kernel size 3x3, a batchnormalization layer~\cite{Ioffe:2015:BNA:3045118.3045167} and a leaky-relu activation.

\subsection{Training}
Our proposed framework consists of two parts: video frame reconstruction and latent code prediction, so the objective function to train both components end-to-end can be formulated as:
\begin{equation}
\begin{split}
    \mathcal{L} =  \lambda_{r}\sum_{q=2}^{k}\sum_{j=1}^N||\hat{T}_{q,j}-T_{q,j}||_2^2 + \\ \lambda_{p}\sum_{m=1}^{M}||\hat{z}_{t,m}-z_{t,m}||_2^2 + \gamma||W||_2^2
\end{split}
\end{equation}

The first term measures the pixel-wise reconstruction loss where $N$ is the total number of pixels per frame and $k$ is the number of frames in a input sequence. We can only reconstruct the last $k-1$ frames if we input $k$ frames since the reconstruction of one frame requires the features from its previous frame. The second term is the MSE between the predicted and the observed latent code where $M$ is the number of elements in the latent code. The last term is an L2 regularization term where $\gamma$ is kept to be 0.001. The model can be trained end-to-end but to bootstrap the encoder with useful features, we first focus on training the autoencoder for reconstruction ($\lambda_{r}=1$ and $\lambda_{p}=0.001$) until the training loss for reconstruction converges. Then we focus on finetuning the weights of the motion model ($\lambda_{r}=0.001$, $\lambda_{p}=1.0$) to use the motion information better.

\subsection{Inference}
At inference time, we discard the decoder and use the difference between predicted and actual latent code as the metric to determine whether a frame is an anomaly or not. The underlying assumption is that the model can predict the latent code for the normal frames with high accuracy but is not able to do so for anomalous frames. To measure the distance between latent codes, we apply two different metrics: Mean Squared Error (MSE) Eq.~\ref{eq:mse} and cosine-distance Eq.~\ref{eq:cosine_dist}. %The cosine-distance metric is considered since it introduces the normalization in the calculation and can be useful if the scene includes many objects with similar size.
\begin{subequations}
\begin{equation}
    s_{t} = \frac{\sum_{m=1}^M ||\hat{z}_{t,m}-z_{t,m}||^2}{M}
\label{eq:mse}
\end{equation}
\begin{equation}
    s_{t} = 1 - \frac{\sum_{m=1}^M \hat{z}_{t,m}z_{t,m}}{\sqrt{\sum_{m=1}^M\hat{z}_{t,m}}\sqrt{\sum_{m=1}^Mz_{t,m}}}
\label{eq:cosine_dist}
\end{equation}
\end{subequations}

After calculating the anomaly score for each frame, following~\cite{Mathieu2015DeepMV}, we normalize the score for each frameset $\mathcal{G}$ to the range of [0,1] using Eq.~\ref{eq:normalize}. A frame that has the anomalous score higher than a threshold is considered as anomaly. Depending on the dataset, $\mathcal{G}$ contains all frames of a video or just the frames in a sliding window for long videos.

\begin{equation}
    s_t = \frac{s_t - min_{j\in\mathcal{G}}(s_j)}{max_{j\in\mathcal{G}}(s_j) - min_{j\in\mathcal{G}}(s_j)}
    \label{eq:normalize}
\end{equation}

\section{Experiments}
\label{sec:experiments}
In this section, we compare our methods to state-of-the-art approaches on public benchmarks. We do not only focus on detection accuracy but also compare the computational cost of the models since this is often the bottleneck that limits the performance in the real world. In addition, we also evaluate the proposed anomaly detector on multiple distorted environments and show that our method is more robust against changes in lighting and weather that are common in the real world.

\subsection{Experimental Setup}
To evaluate the effectiveness of our proposed methods, we use the same datasets as~\cite{liu2018ano_pred}, including the UCSD Pedestrian dataset~\cite{Mahadevan.anomaly.2010}, the CUHK Avenue dataset~\cite{Lu2013AbnormalED} and the ShanghaiTech dataset~\cite{liu2018ano_pred}. The details of these datasets are described by~\cite{liu2018ano_pred}. As is common, we report the Area Under Curve (AUC) score as the accuracy metric. 

The experimental settings are shown in Table \ref{tab:implementation_detail}. We use 8 input frames for UCSDPed1 and UCSDPed2 datasets and 6 input frames for the Avenue and ShanghaiTech datasets. More frames are needed to encode the motion and appearance of the much smaller objects in the UCSD Pedestrian datasets than in the Avenue and ShanghaiTech datasets. The resolution of the frames is reduced using bilinear interpolation, keeping the original aspect ratio. As for the architecture, we use five encoder blocks for the UCSDPed1 dataset and four encoder blocks for the other datasets. This is done because the scenes in the UCSDPed1 dataset include more objects and we need to increase the model's capacity in order to encode the appearance- and motion-related features. 
\begin{table}[ht!]
\caption{Design choices for each evaluation dataset. }
\resizebox{0.76\textwidth}{!}
{\begin{minipage}{\textwidth}
\begin{tabular}{ccccc}
\toprule
 & UCSDPed1 & UCSDPed2 & Avenue & ShanghaiTech \\ \toprule
Input Size & 128x192 & 128x192 & 128x224 & 128x224 \\
Num Input ($k$) & 8 & 8 & 6 & 6 \\
Encoder Block & 5 & 4 & 4 & 4 \\ \bottomrule
\end{tabular}
\end{minipage}}
\label{tab:implementation_detail}
\end{table}

We train the model for 50 epochs in an end-to-end fashion with initial learning rate $1e-4$ which decays by 0.1 every 20 epochs. %In the first 25 epochs, we focus on training the autoencoder with learning rate $1e-4$. Then, we put more attention on learning the motion model in the next 25 epochs with an initial learning rate $1e-4$, which decays by 0.1 every 20 epochs.
The ShanghaiTech dataset contains data from multiple cameras. We trained individual models per camera and observed no significant performance difference with a model that is trained on data from all cameras. We train the model with the Adam optimizer~\cite{DBLP:journals/corr/KingmaB14} for all our experiments. The code will be released on \textbf{website}.

For the evaluation, we use feature-wise MSE (Eq.~\ref{eq:mse}) to calculate the anomaly score for Avenue dataset and feature-wise cosine-distance (Eq.~\ref{eq:cosine_dist}) for all other datasets.
%We explain the reason for doing this in Section. 
We normalize the anomaly score in UCSDPed1 and UCSDPed2 dataset with Eq.~\ref{eq:normalize} using all the frames in a test video. For the ShanghaiTech datasets, we use the same sliding-window approach as~\cite{abati2019latent}. %, and $\mathcal{G}$ only consists of the $m$ closest frames. In our experiment, $m$ is set to be 200. 
%Additionally, we do not normalize the anomaly score in Avenue dataset since the high prediction error for the distortion due to the camera movements on the second test video will omit the recognition of other correctly predicted anomalies.  

\subsection{Results}
\label{sec:exp_result_auc_fps}
Table \ref{tab:experiment_result_auc} compares our results with those of other unsupervised deep learning based methods for anomaly detection. Our approach outperforms the existing methods in terms of frame-level AUC score on all four datasets. The decoupled mechanism and the combined learning of appearance and motion information improves the training process and allows the extracted and predicted latent codes to be more representative for the frequently seen events and thus improve the anomaly detection accuracy. We further investigate the role of decoupling, the use of Conv3D layers in the motion model as well as different anomaly metrics in section~\ref{sec:ablation_study}.
\begin{table}[ht!]
\caption{Frame-level AUC score with 95\% confidence interval (4 runs) on UCSDPed1, UCSDPed2, Avenue and ShanghaiTech datasets. We outperform most of the existing approaches on all the datasets.} \smallskip
\resizebox{0.89\textwidth}{!}
{\begin{minipage}{\textwidth}
\begin{tabular}{@{}lcccc@{}}
\toprule
 & \begin{tabular}[c]{@{}l@{}}UCSD\\ Ped1\end{tabular} & \begin{tabular}[c]{@{}l@{}}UCSD\\ Ped2\end{tabular} & Avenue & \begin{tabular}[c]{@{}l@{}}Shanghai\\ -Tech\end{tabular}\\ \midrule
\begin{tabular}[c]{@{}l@{}}MDT \cite{5539872}\end{tabular} & 81.8 & 82.9 & - & -  \\ \noalign{\smallskip}
\begin{tabular}[c]{@{}l@{}}ConvAE \cite{Hasan2016LearningTR}\end{tabular} & 81.0 & 90.0 & 70.2 & - \\ \noalign{\smallskip}
\begin{tabular}[c]{@{}l@{}}ConvLSTM \cite{8019325}\end{tabular} & 75.5 & 88.1 & 77.0 & -  \\ \noalign{\smallskip}

\begin{tabular}[c]{@{}l@{}}Unmasking \cite{Ionescu2017UnmaskingTA}\end{tabular} & 68.4 & 82.2 & 80.6 & -\\ \noalign{\smallskip}

\begin{tabular}[c]{@{}l@{}}Hinami \cite{hinami2017joint}\end{tabular} & - & 92.2 & - & - \\ \noalign{\smallskip}

\begin{tabular}[c]{@{}l@{}}StackRNN \cite{8237307}\end{tabular} & - & 92.2 & 81.7 & -  \\ \noalign{\smallskip}

\begin{tabular}[c]{@{}l@{}}FFP-MC \cite{liu2018ano_pred}\end{tabular} & 83.1 & \textbf{95.4} & 84.9 & 72.8 \\ \noalign{\smallskip}

\begin{tabular}[c]{@{}l@{}}LatentAuto 
\cite{abati2019latent}\end{tabular} & - & \textbf{95.4} & - & 72.5  \\ \hline \hline
ours & \textbf{84.7$\pm$0.3} & \textbf{95.1$\pm$0.3} & \textbf{88.8$\pm$0.3} & \textbf{74.2$\pm$0.1}\\ \bottomrule
%ours FPS & 81 & 104 & 79 & \\ \bottomrule
\end{tabular}
\end{minipage}}
\label{tab:experiment_result_auc}
\end{table}

%Note, we did not compare our results with the state-of-art from Ionescu \etal~\cite{Ionescu_2019_CVPR} because: i) Ionescu \etal apply a supervised object detector before the anomaly detection whereas our method is fully unsupervised ii) The AUC score reported by Ionescu \etal is the average AUC score over all the testing videos while we calculate the AUC score based on the frames from all the testing videos for each dataset. 

The initial goal of our approach was to develop an architecture that is more efficient than previous approaches.
We benchmark our method with the approaches that have publicly available code on an Intel(R) Core(TM) i7-8700 CPU @ 3.20GHz with an GeForce GTX 1080 Ti. We reimplemented the methods described by~\cite{Hasan2016LearningTR} in Tensorflow to allow for a fair comparison\footnote{The benchmark code is also available at \textbf{website}}. The results are displayed in Table~\ref{tab:experiment_result_fps}. %The anomaly detection results using method \textit{LatentAuto} on UCSDPed1 and Avenue datasets are not provided in~\cite{abati2019latent}, so we simply assume the FPS for UCSDPed1 is as same as UCSDPed2 and the FPS for Avenue is as same as ShanghaiTech.
Our method outperforms the other approaches by a large margin in terms of the number of frames that can be processed per second. Compared to the \textit{Latent-Auto} approach~\cite{abati2019latent} that also detects anomalies using latent code, we can process 16$\sim$45 times more frames per second. The main reason our method is more efficient is because the model independently encodes appearance and motion information. We extract the appearance information using a  relatively efficient 2D convolutional network and process the combined features using a small 3D convolutional network, whereas other approaches build the entire network around 3D convolutions making it much more expensive. Because we extract latent codes from individual frames, each frame only needs to be processed once. We can save each of the last $k$ latent codes that are needed in the motion model and re-use them for the next $k$ predictions. % uses the last $k$ latent codes but each of these latent codes only need to % The motion model combines features from the last $k$ frames but each of these features are only extracted once and then kept in memory to be reused for the other $k-1$ predictions. 
In contrast, models that use 3D convolutions process each frame $k$ times, each time at a different position in the stack, predicting a different frame, making it much more computationally expensive. Since we predict future latent codes and use the prediction error in latent space as our anomaly metric, we do not need the decoder part at inference, again reducing the computational cost. Also, compared to other models that use anomaly scoring metrics based on latent codes, we do not impose any distribution constraint on the latent code, giving the model the freedom to fit the data as best as possible.
%However, we can process less frames than ConvAE method~\cite{Hasan2016LearningTR} on a CPU device. One possible reasons is that the 3D-Convolutional layers are more optimized for the GPU rather than the CPU~\cite{conv3d_github}. 
\begin{table}[ht!]
\caption{FPS for different methods. Our method is more efficient than the existing approaches}
\resizebox{0.8\textwidth}{!}
{\begin{minipage}{\textwidth}
\begin{tabular}{@{}lllll@{}}
\toprule
 & \begin{tabular}[c]{@{}l@{}}ConvAE  \cite{Hasan2016LearningTR}\end{tabular} & \begin{tabular}[c]{@{}l@{}}FFP+MC \cite{liu2018ano_pred}\end{tabular} & 
 \begin{tabular}[c]{@{}l@{}}LatentAuto \cite{abati2019latent}\end{tabular}
 & Ours \\ \midrule
%CPU & 0.048s(20) & 0.260s(3) & 51.022(0.02)  & 0.110s(9) \\
UCSDPed1 & 75 & 63 & 2 & \textbf{81} \\
UCSDPed2 & 75 & 63 & 2 & \textbf{90} \\ 
Avenue & 71 & 48 & 5 & \textbf{77} \\
ShanghaiTech & 71 & 48 & 5 &\textbf{77} \\
\bottomrule
%AUC & 88.8 & 95.4 & 95.4 & 95.4 \\ \bottomrule
\end{tabular}
\end{minipage}}
\label{tab:experiment_result_fps}
\end{table}

\begin{figure*}[ht!]
\centering
\resizebox{1.0\textwidth}{!}
{
\begin{minipage}{\textwidth}
\includegraphics[width = 1.0\textwidth]{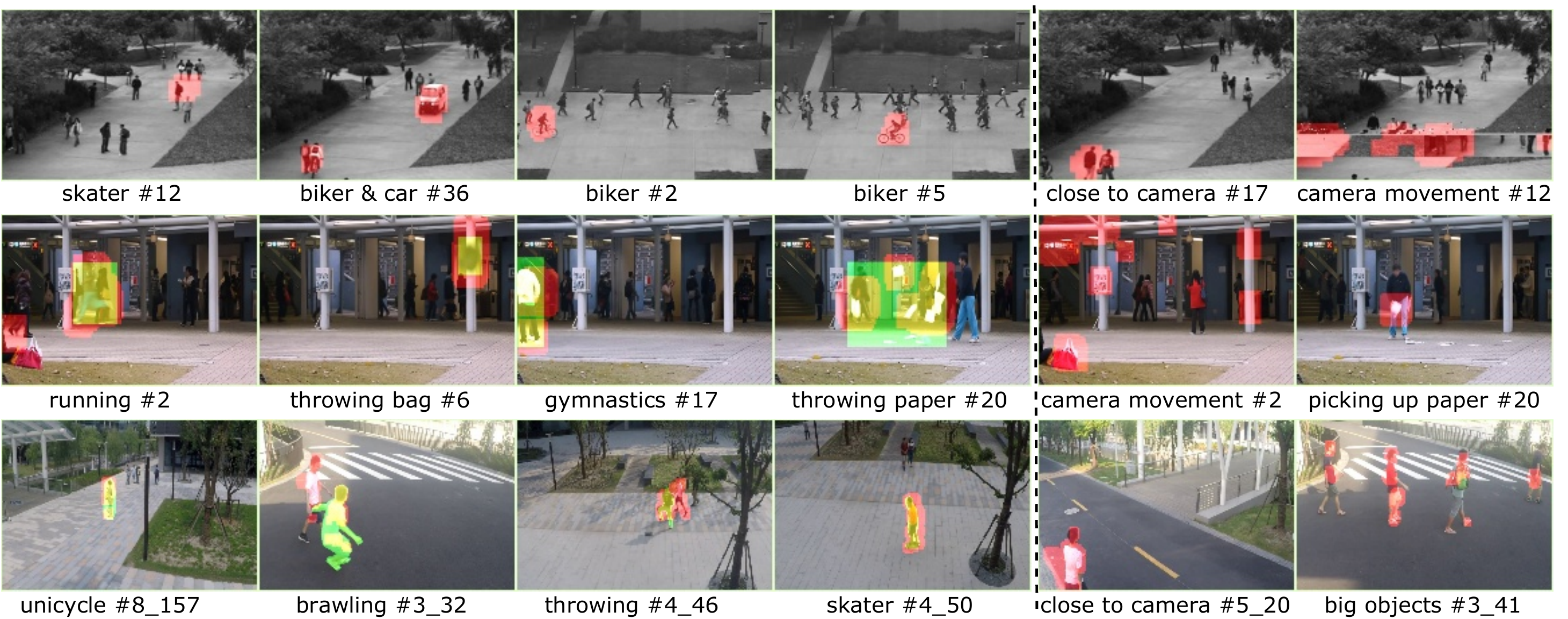}
\end{minipage}
}
\smallskip
\caption{True positive (first four columns) and false positive (last two columns) detections of our framework. Examples are selected from the UCSD pedestrian dataset (first row), Avenue dataset (second row) and ShanghaiTech dataset (last row). \# indicates the testing video index. The red boxes are the regions that have highest pixel-wise prediction error and the green box are the ground truth bounding boxes for the anomalous event. We can successfully detect the motion- and appearance- related anomalies and tend to incorrectly flag objects as anomalies if they are closer to camera. The figure is best viewed in color.}
\label{fig:qualitative_result}
\end{figure*}

A disadvantage of working with latent codes is that it is harder to interpret the model. It is however possible to also use the decoder at inference time to generate predicted frames and to measure the pixel wise reconstruction metrics. This allows us to localize the part of the frame that contains the anomaly. We show these results in Figure \ref{fig:qualitative_result}. The red boxes in each frame are the regions that have the prediction error larger than a threshold.  The green boxes show the ground-truth annotations for the Avenue and ShanghaiTech dataset (the UCSD pedestrian dataset only has frame-level labels). These results empirically confirm that our model can detect motion- and appearance-related anomalies, such as the \textit{skater, cyclist, car, running and gymnastics} events (first four columns). The last two columns of Figure \ref{fig:qualitative_result} show false positives, frames that were labeled as normal but that were flagged as anomalies by our model. Two of these show noise or camera movements that were not seen during training. It also shows that the model is more likely to incorrectly flag objects as anomalies if they are closer to the camera.

%5. Yes, I also need to explain that I am using the sliding window for shanghaitech. No, i will first wait for the shanghaitech dataset, cuz i need to train the model for each background. 

% That's all for the result. 

% Chong \etal~\cite{10.1007/978-3-319-59081-3_23} use the GPU (NVIDIA Maxwell Titan X) \\
% Liu \etal~\cite{liu2018ano_pred} use the NVIDIA GeForce TITAN GPUs with Intel Xeon(R) E5-2643 3.40GHz CPUs
% and Samsung SSD 850 PRO. \\
% I am using GeForce GTX 1080 Ti

\section{Robustness of the model}
\label{sec:robustness_of_the_model}
The datasets we used in the previous section are commonly used datasets that allow us to compare anomaly detection techniques quantitatively. They however all contain relatively clean data, recorded at similar times during the day and under clear weather conditions. These datasets are therefore not necessarily representative of real world surveillance footage where external factors such as weather and time of day will severely influence the performance of the model. We argue that in addition to their anomaly detection performance and computational cost, we should also compare the robustness and generalization of the models to these external factors. In this section we investigate two types of robustness and show that by working with latent code anomaly metrics we are more robust than other approaches that use pixel-wise metrics.

\subsection{Modelling long term temporal information}
In section~\ref{sec:exp_result_auc_fps}, we showed that our approach is by design much more efficient than existing techniques. To reduce the computational cost even further we could reduce the number of times we activate the model. In surveillance video, anomalies are typically in view of the camera during multiple seconds. It should be enough to process only a few frames of this window to detect the anomaly. If instead of running our model every 40 ms, we run it every 200 ms, then this obviously results in a lower total computational cost but this also makes the task much harder for the network since we now need to predict five times further into the future. In this way, the model is forced to encode longer term temporal information and the prediction task is more challenging since the future frame will differ substantially from the previous frames.

%The behavior of the anomalous object is usually continuous in the video, so the identification of the anomaly will last until it disappears. It indicates that probably not all the frames need to be evaluated to flag an abnormal event as long as we can capture any one of the frames during this event. To simulate this situation, we increase the stride between the input frames as well as the stride between input and output frames. In this way, the model is forced to encode the long term temporal information and the decision for a much further frame can be made without evaluating the intermediate frames. 

\begin{figure}[ht!]
    \centering
    \includegraphics[width = .45\textwidth]{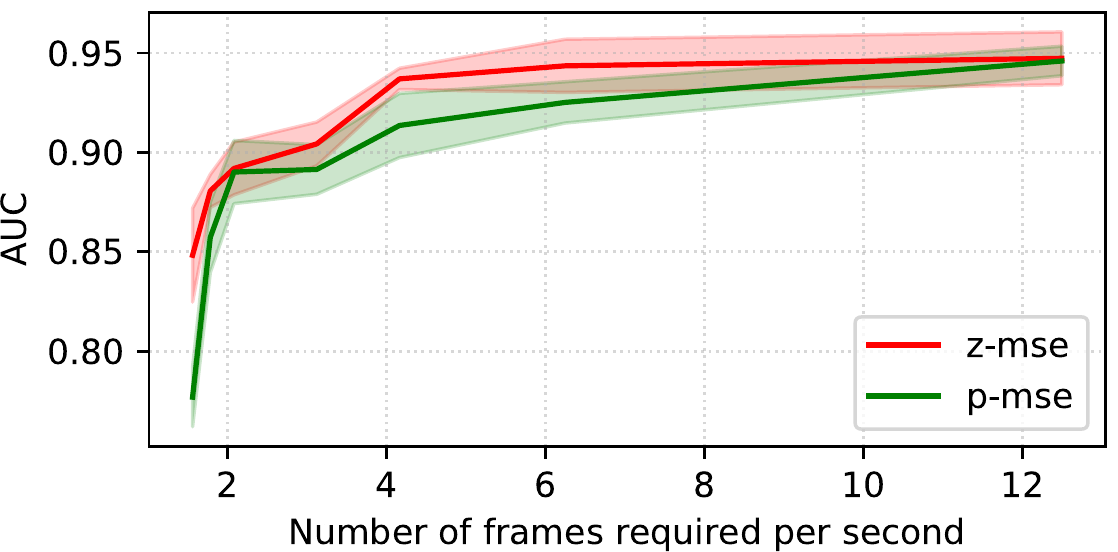}
    \caption{The anomaly detection accuracy (frame-level AUC score with 95\% confidence interval) when we have low fps input. \textbf{z-mse} is calculated using Eq.~\ref{eq:mse} on the latent code and \textbf{p-mse} is the MSE between the predicted frames and actual frames. The figure is best viewed in color.}
    \label{fig:low_fps_input}
\end{figure}

To explore this trade-off, we subsample the video sequence and only keep every $d_{th}$ frame in our training and test data. Figure~\ref{fig:low_fps_input} shows how sensitive the latent code metric (red line) is compared to the pixel-wise metric (green line). Both techniques follow the same trend but the latent code metric consistently performs better than the pixel-wise metric, especially when the gap between input frames becomes large. This illustrates the power of latent codes and their capability of modelling longer term temporal information.

\subsection{Generalization to other lightning conditions}
\begin{figure}[t]
    \centering
    \includegraphics[width=.47\textwidth]{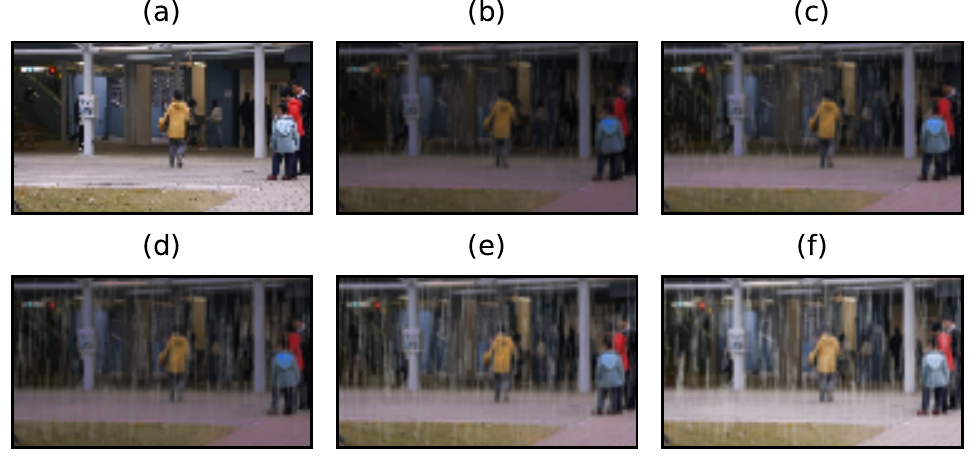}
    \caption{The distorted frames using Avenue dataset. (a) original frame,  (b) and (c) have heavy rain with brightness degree 0.5 and 0.7 respectively and (d),(e) and (f) show torrential rain with the brightness 0.6, 0.8 and 1.0 respectively. The figure is best viewed in color.}
    \label{fig:avenue_augmented_frame}
\end{figure}

The performance of anomaly detection in surveillance video is impacted severely by factors such as varying illumination, multiple weather conditions, on- and off-peak traffic profiles, degradation of the camera and so on. Therefore, in this section, we investigate the robustness of our proposed method to these distortions. We train a model using original frames from the Avenue dataset and then analyze the performance on distorted test set frames. We adjust brightness, blur the image and add rain to the test frames using the Automold toolkit\footnote{\url{https://github.com/UjjwalSaxena/Automold--Road-Augmentation-Library}}. Figure \ref{fig:avenue_augmented_frame} shows some examples of the distorted frames using different levels of rain and brightness.

\begin{figure}[ht!]
    \centering
    \includegraphics[width=.42\textwidth]{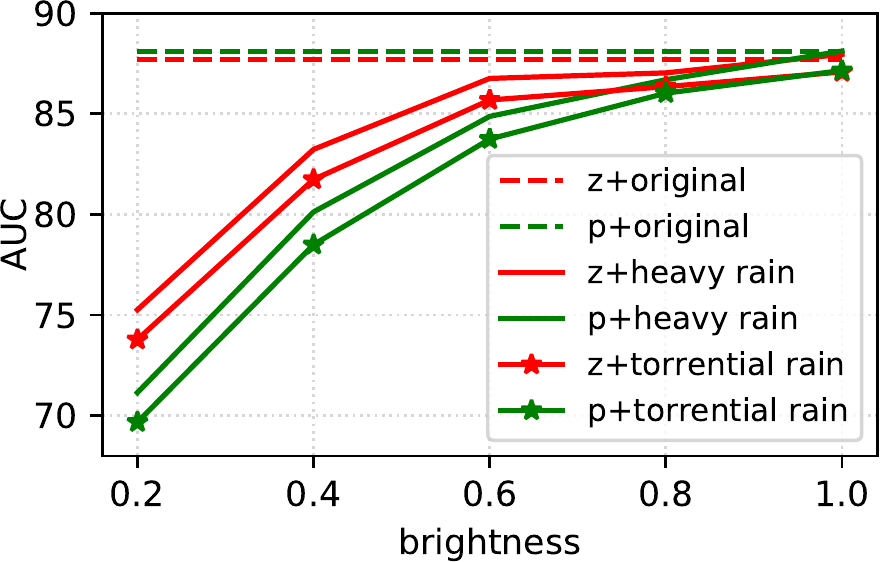}
    \caption{The averaged anomaly detection accuracy (frame-level AUC score) on the augmented frames where \textbf{z} means the latent code feature-wise MSE and \textbf{p} means the prediction pixel-wise MSE. The latent code anomalous score measurement is more robust on unseen weather conditions compared to other approaches. The figure is best viewed in color.}
    \label{fig:avenue_exp_on_augmented_frame}
\end{figure}

Figure \ref{fig:avenue_exp_on_augmented_frame} shows the frame-level anomaly detection accuracy for different distortion levels. The X-axis shows the relative brightness compared to the original frame. The different curves show the performance of using Mean Squared Error (MSE) in pixel space (p) and in latent space (z) as our anomaly metrics for different levels of rain added to the image. As expected the model performance drops when the brightness decreases, but our model is consistently more robust than the baseline model that uses pixel wise metrics. Adding rain to the test frames also reduces the detection performance  but again, our feature-wise latent code MSE performs significantly better as anomaly scoring metric than the pixel-wise MSE. These results verify that our proposed methods using feature-wise MSE in the latent code to identify anomalies is more robust to different outdoor situations than pixel-wise MSE measurements.

\section{Ablation study}
\label{sec:ablation_study}
In section \ref{sec:method} we introduced our model together with some design choices. In this section we look closer to three of these and investigate how these contribute to our results.

\subsection{Reconstruction vs prediction}
To understand the impact of the anomaly detection metrics on the detection accuracy, we report the frame level AUC score using pixel-wise reconstruction error, pixel-wise prediction error and feature-wise latent code error in Table~\ref{tab:ucsd_avenue_shanghai_quan_motion}. Adding the motion model highly improves the anomaly detection accuracy for all the datasets since it encodes spatiotemporal information better. Compared to the performance using pixel-wise prediction error, the use of latent code prediction error tends to be better and more stable since it is more robust to the noise in the image as indicated by section~\ref{sec:robustness_of_the_model}. To qualitatively understand how the model differentiates between normal and abnormal frames, we conduct an experiment on the moving-mnist dataset.

\begin{figure}[ht!]
    \centering
    \includegraphics[width=.47\textwidth]{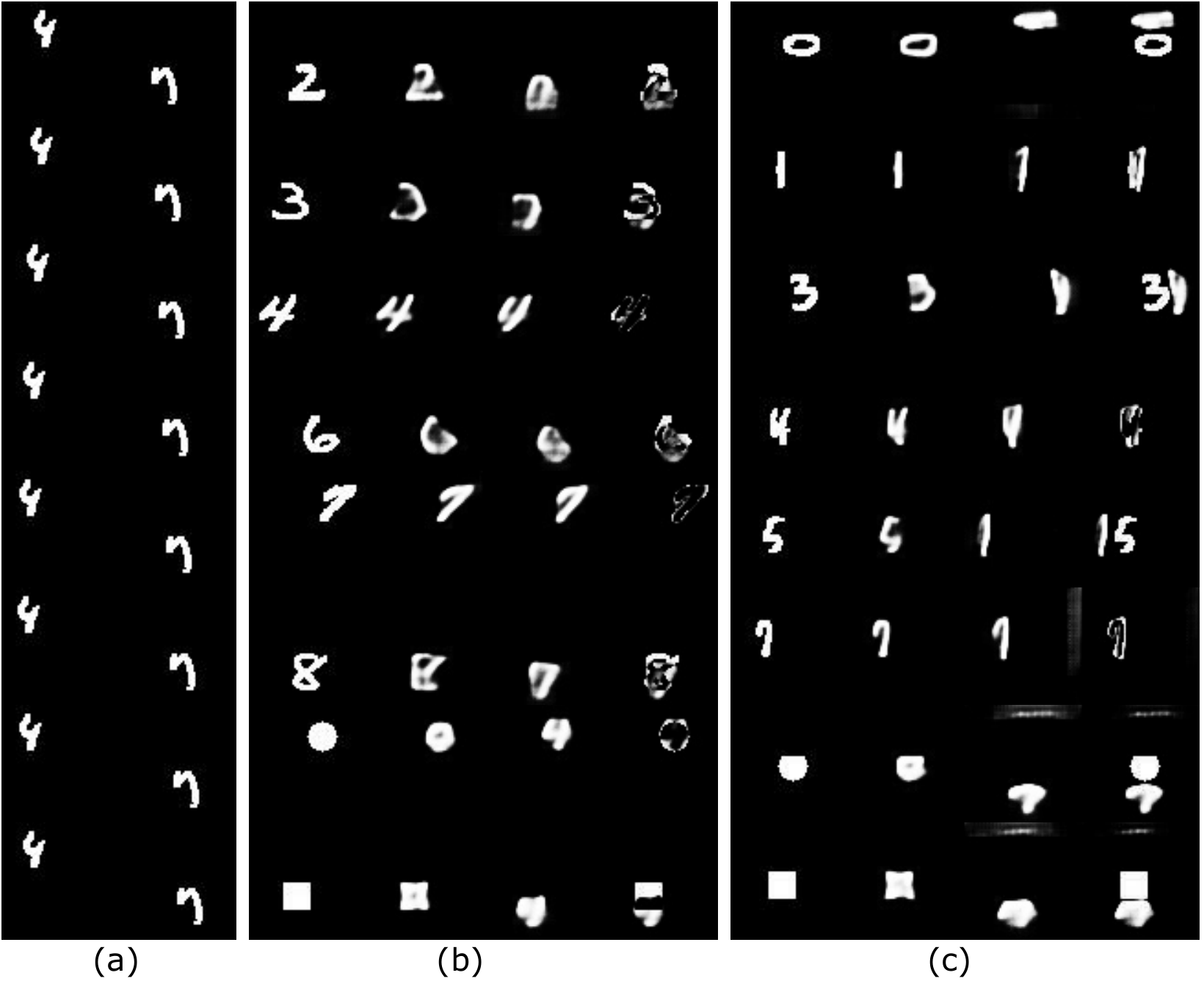}
    \caption{Experimental results on moving-mnist dataset. (a) training digits 4 and 7 are moving horizontally or vertically randomly with speed 2. (b) and (c) show the testing digits and shapes that are moving in a similar fashion but with speed 2 (b) and speed 4 (c). From left to right, the columns in (b) and (c) are input, reconstruction, prediction and prediction error. The model cannot accurately reconstruct and predict for both appearance-related (unseen digits) and motion-related (unseen moving speed) anomalies.}
    \label{fig:ucsd_avenue_to_explain_latent_space}
\end{figure}

We train the same model as shown in Fig.~\ref{fig:main_architecture} using the video sequences that are created by letting randomly selected digits 4 and 7 from the training set of MNIST~\cite{726791} move horizontally or vertically with a speed of 2 following~\cite{8237307} (see Fig.~\ref{fig:ucsd_avenue_to_explain_latent_space} (a)). This model is then tested on video sequences that include all types of digits from the test set of MNIST dataset and two new shapes (circle and square) that are moving also horizontally or vertically with a speed of 2 or speed of 4. The input, reconstruction, prediction and prediction error are shown in Fig.~\ref{fig:ucsd_avenue_to_explain_latent_space} (b) and (c).

Fig.~\ref{fig:ucsd_avenue_to_explain_latent_space} (b) shows the model output for input objects that move with a speed of 2. The model can make good reconstructions and predictions for already seen digits 4 and 7 but tend to predict the unseen digits to be one of the already seen digits. For example, it predicts the circle and square to be similar to 4 and digit 8 to be similar to 7. The difference between the reconstruction and prediction indicates that the model cannot make a good prediction of the latent code for the unseen digits and this allows us to detect the appearance related anomalies. For the objects that are moving with a higher speed as shown in Fig.~\ref{fig:ucsd_avenue_to_explain_latent_space} (c), the model produces a prediction that falls behind the actual input. This is because the designed motion model can further exploit the speed mismatch of the objects during training and testing and is thus able to detect the motion related anomalies.

\begin{table*}[ht!]
\caption{Abnormal event detection results (in \%) with 95\% confidence interval. The \textit{Reconstruction error} and \textit{Prediction error} are pixel-wise MSE and the \textit{Latent code error} is calculated feature-wise. We achieved better performance using latent-code prediction error as anomalous score and using Conv3D layers in the motion model.}
\label{tab:ucsd_avenue_shanghai_quan_motion}
\centering
\resizebox{0.95\textwidth}{!}
{\begin{minipage}{\textwidth}
\begin{tabular}{@{}lcccccccc@{}}
\toprule
                     & \multicolumn{4}{c}{Conv3D}          & \multicolumn{4}{c}{ConvLSTM}        \\  \midrule
                     & Ped1 & Ped2 & Avenue & ShanghaiTech & Ped1 & Ped2 & Avenue & ShanghaiTech \\ \cmidrule(l{1.5mm}r{3mm}){2-5} \cmidrule(l{1.5mm}r{3mm}){6-9}
Reconstruction error & 80.6$\pm$0.2 & 91.3$\pm$0.5 & 83.7$\pm$0.3 &  55.8$\pm$0.2 & 74.8$\pm$0.4 & 87.5$\pm$0.1 &83.1$\pm$0.2 &  52.6$\pm$0.4\\
Prediction error     & 82.2$\pm$0.4 & 90.9$\pm$0.5 & \textbf{89.2$\pm$0.2} &  72.7$\pm$0.2 & 81.2$\pm$0.4& 90.3$\pm$0.8 & \textbf{89.0$\pm$0.3}&     70.2$\pm$0.2          \\
Latent code error    & \textbf{84.7$\pm$0.3} &\textbf{95.1$\pm$0.3}& \textbf{88.8$\pm$0.3} & \textbf{74.2$\pm$0.1} & 82.8$\pm$0.6 & 94.1$\pm$0.1 &   \textbf{88.9$\pm$0.3}     &    71.3$\pm$0.4\\ \bottomrule
\end{tabular}
\end{minipage}}
\end{table*}

\subsection{Design of the motion model}
\label{sec:design_motion_model}
The motion model is an important factor in the design of our architecture since it is required to encode the typical appearance and motion information of the frequently seen events. Therefore, we also replaced the Conv3D layers in the motion model with ConvLSTM to study the impact of the design of the motion model on anomaly detection performance. The anomaly detection accuracy using pixel-wise reconstruction, prediction error and feature-wise latent code error for different datasets are shown in Table~\ref{tab:ucsd_avenue_shanghai_quan_motion}. We achieved better or similar performance on all the datasets using Conv3D layers in the motion model. One of the possible reasons is that the Conv3D layers can fit the data better and thus extract more representative features. In addition, we observed that the model that uses ConvLSTM layers have delayed detection results such that it fails to detect the beginning of anomalous events and it also reports more false alarms after the anomalous events due to the slowly response.

\section{Conclusion and future work}
\label{sec:conclusion}
In this paper we introduced a novel architecture that is able to detect anomalies in real world surveillance footage using only unsupervised training. The model consists of two parts where the first part extracts appearance features from individual frames and the second part uses these features to predict the latent code for a future frame. In contrast to previous works,  our model uses a prediction in latent space as a metric to detect anomalies. We showed that is able to outperform other techniques that use reconstruction or pixel based prediction metrics. Because of the decoupled appearance and motion feature learning, our model is also much more efficient than related approaches. Where other techniques use expensive 3D convolutions to analyze a stack of frames, we process each frame individually and then combine the information with a much smaller 3D convolutional model. This allows us to process 16 to 45 times more frames using the same computational budget. Finally, we show that using latent space features makes the model more robust against distortions such as changing lighting or weather conditions.

Anomaly detection in real world surveillance data is a very challenging topic with many useful applications. For future work, we argue that more research is needed to deal with changing environments, weather and lighting conditions as well as with camera degradation. 
\section*{Acknowledgments}
This research received funding from the Flemish Government under the “Onderzoeksprogramma Artificiële Intelligentie (AI) Vlaanderen” programme, and from imec under the CityFlows AAA programme.

%\begin{table}
%\begin{center}
%\begin{tabular}{|l|c|}
%\hline
%Method & Frobnability \\
%\hline\hline
%Theirs & Frumpy \\
%Yours & Frobbly \\
%Ours & Makes one's heart Frob\\
%\hline
%\end{tabular}
%\end{center}
%\caption{Results.   Ours is better.}
%\end{table}

%\begin{figure}[t]
%\begin{center}
   %\includegraphics[width=0.8\linewidth]{egfigure.eps}
%\end{center}
%   \caption{Example of caption.  It is set in Roman so that mathematics
%   (always set in Roman: $B \sin A = A \sin B$) may be included without an
%   ugly clash.}
%\label{fig:long}
%\label{fig:onecol}
%\end{figure}

%\begin{figure*}
%\begin{center}
%\fbox{\rule{0pt}{2in} \rule{.9\linewidth}{0pt}}
%\end{center}
%   \caption{Example of a short caption, which should be centered.}
%\label{fig:short}
%\end{figure*}

{\small
\bibliographystyle{ieee_fullname}
\bibliography{egbib}
}

\end{document}